\documentclass[runningheads]{llncs}
\usepackage[T1]{fontenc}
\usepackage{graphicx,verbatim}
\usepackage{amsmath,amssymb}
\usepackage{booktabs}
\usepackage[table]{xcolor}
\begin{document}

\title{P\textsuperscript{3}CA: Encoder-Agnostic Interpretation of\\Vision Foundation Model Embeddings via\\Spatial Probing}

\titlerunning{P\textsuperscript{3}CA for Spatial Representation Probing}
% If the paper title is too long for the running head, you can set
% an abbreviated paper title here
%
\author{
Amoon Jamzad\inst{1} \and
Dilakshan Srikanthan\inst{1} \and
Faranak Akbarifar\inst{1} \and
Nooshin Maghsoodi\inst{1} \and 
Parvin Mousavi\inst{1}
}
\authorrunning{A. Jamzad et al.}
% First names are abbreviated in the running head.
% If there are more than two authors, 'et al.' is used.
%
\institute{School of Computing, Queen's University, Kingston, Ontario, K7L 2N8, Canada \\
\email{a.jamzad@queensu.ca}}

\maketitle              % typeset the header of the contribution

\begin{abstract}
Vision foundation models are increasingly used as reusable encoders in medical image computing, yet their high-dimensional spatial embeddings are difficult to inspect beyond downstream task performance or global dimensionality reduction. We propose position-prompted PCA (P\textsuperscript{3}CA), an encoder-agnostic method for local probing of channel-rich spatial tensors. Given a user-selected spatial prompt, P\textsuperscript{3}CA estimates the feature normalization and dominant covariance directions within that region, then applies the resulting projection to the full tensor to visualize where locally informative directions are expressed. This produces a region-conditioned representation lens without modifying the encoder, retraining, or requiring task-specific labels. We implement P\textsuperscript{3}CA in EmbedVision, an interactive 3D Slicer-based workflow, and evaluate it across natural images, colorectal pathology foundation-model embeddings, and spatial transcriptomic tensors. Across these settings, prompted projections reveal local structure suppressed by global PCA, improve prompt-matched pathology discrimination from frozen three-dimensional projections, and support comparison between learned and measured spatial representations.
\end{abstract}

\section{Introduction}

Vision foundation models are increasingly used as general-purpose encoders in medical image computing, providing transferable representations for diverse downstream tasks including classification, segmentation, retrieval, registration, and visual question answering~\cite{ma2024medsam,zhang2024foundation}. By producing high-dimensional feature maps, these encoders can reduce the need for task-specific training~\cite{oquab2023dinov2,simeoni2025dinov3}. However, medical images contain clinically distinct regions, such as lesions, healthy tissue, anatomical boundaries, and artifacts, and strong benchmark performance does not guarantee that models represent these distinctions rather than relying on shortcuts~\cite{jimenez2023shortcuts}. This motivates methods that go beyond downstream evaluation to inspect how spatial representations are organized across image regions.

These representations can be viewed as channel-rich spatial tensors, where each location contains a high-dimensional latent feature vector~\cite{oquab2023dinov2,simeoni2025dinov3}. The same view extends to mass spectrometry imaging, hyperspectral imaging, and spatial omics, where channels correspond to molecular, spectral, or gene-expression measurements rather than learned embeddings~\cite{massvision,lu2014medicalhsi,bressan2023spatialomics}. This shared structure motivates a unified framework for region-aware exploration of high-dimensional spatial data, focused on vision foundation model embeddings but applicable to measured biomedical modalities.

Existing evaluation and interpretation methods offer useful but incomplete views of spatial representations. Extrinsic approaches, such as probes, prediction heads, and downstream benchmarks, measure task utility but require labels and do not show how information is organized within the feature map~\cite{zhan2024probing,elbanani2024probing}. Attribution and saliency methods explain model outputs rather than the representation itself~\cite{selvaraju2017gradcam,saporta2022saliency}. Label-free visualization methods such as PCA, t-SNE~\cite{van2008tsne}, and UMAP~\cite{mcinnes2018umap} can inspect features directly, but their global projections may obscure locally meaningful structure in spatially heterogeneous images. This limitation is especially relevant for medical image computing, where local context often determines interpretation~\cite{chaitanya2020globalLocal}. The directions that are meaningful within tumor tissue may differ from those that separate normal tissue, anatomical boundaries, or artifacts, and such local structure can be obscured by a globally estimated projection. An interpretable representation-analysis method should therefore allow users to ask: what directions dominate variation in this region, how does this local structure differ from the global representation, and where else in the image do similar directions appear?

In this paper, we propose position-prompted PCA (P\textsuperscript{3}CA), a label-free method for locally probing high-dimensional spatial feature maps. Given a user-selected region, P\textsuperscript{3}CA estimates the feature directions that maximize local contrast within that region and applies the resulting projection to the full map, revealing where region-relevant representation structure appears across the image. We implement this workflow in EmbedVision, an open-source interactive framework, and demonstrate it across natural images, histopathology embeddings, and spatial transcriptomic measurements.

Our contributions are threefold. First, we formulate spatial representation probing as a region-conditioned covariance analysis problem over channel-rich spatial tensors. Second, we introduce P\textsuperscript{3}CA as a simple, encoder-agnostic projection method that complements downstream probes, attribution methods, and global embedding visualizations without model modification or retraining. Third, we present an interactive implementation and case-study evaluation showing how spatial prompts support local representation inspection across learned embeddings and measured biomedical tensors.

\section{Method}

\begin{figure}[!t]
\centering
\includegraphics[width=0.9\textwidth]{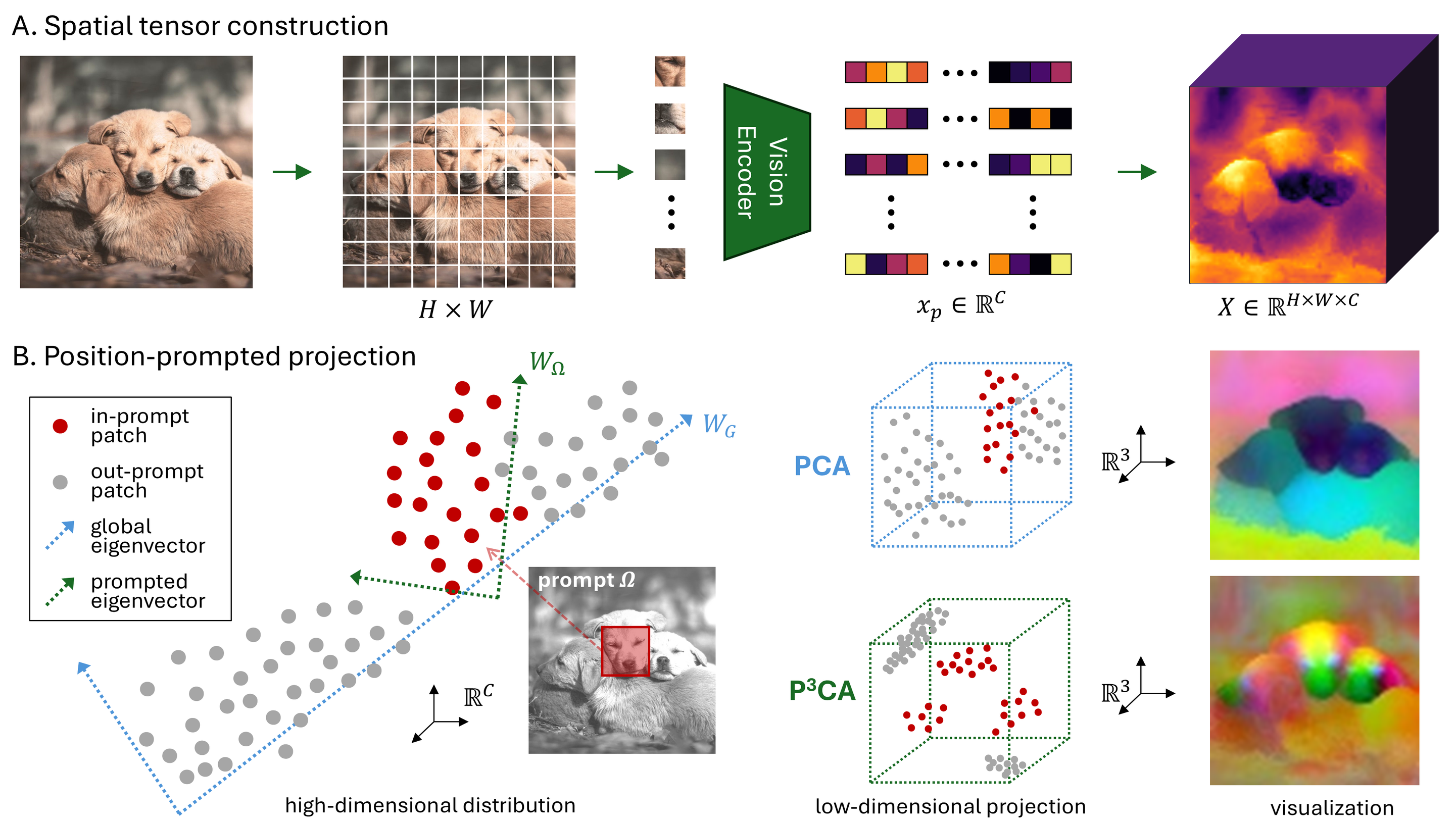}
\caption{P\textsuperscript{3}CA workflow. (A) An image is patchified, encoded, and reassembled into a channel-rich feature tensor. (B) While global PCA estimates basis $W_G$ from all patches, P\textsuperscript{3}CA estimates $W_\Omega$ from a spatial prompt $\Omega$ and applies it to the full tensor to generate an RGB representation map.}
\label{fig:workflow}
\end{figure}

Figure~\ref{fig:workflow} summarizes the workflow. An encoder output is represented as a tensor. P\textsuperscript{3}CA fits a projection basis inside a user-selected spatial prompt and applies the resulting transform to the full tensor, producing a full-field visualization of locally dominant feature directions.

\subsection{Tensor representation}

Given an input image $I$, it is patchified into spatial units, which are mapped by a vision encoder to feature vectors and reassembled on their spatial grid to form
$X \in \mathbb{R}^{H\times W\times C}$,
where $H$ and $W$ define the spatial resolution of the feature map and $C$ is the feature dimension. Each spatial location $p=(i,j)$ is associated with a feature vector $x_p \in \mathbb{R}^{C}$. For learned representations, the channel axis indexes latent embedding dimensions while preserving spatial correspondence with image regions. For measured biomedical modalities, the same tensor notation applies, with channels corresponding to ion intensities, spectra, or gene-expression measurements rather than learned features.

\subsection{Position-prompted PCA}

The goal of P\textsuperscript{3}CA is to generate a three-channel representation map whose contrast is optimized for a user-selected spatial context. Rather than choosing projection directions from the full image, we ask which feature directions vary most within a prompted region and then use those directions as a lens for the full tensor. Let $\mathcal{G}$ denote the full spatial grid and let $\Omega \subset \mathcal{G}$ denote the spatial prompt. We first fit a feature-wise preprocessing transform $T_{\Omega}$ using only vectors inside $\Omega$ and apply it to all spatial locations, giving $u_p = T_{\Omega}(x_p)$ for $p \in \mathcal{G}$, where $T_{\Omega}$ includes normalization and centering using prompt-region statistics.

We define local contrast as the variance of projected feature values within the prompted region. For a unit projection direction $w \in \mathbb{R}^{C}$, this objective can be written as
\begin{equation}
C_{\Omega}(w)
=
\frac{1}{|\Omega|-1}
\sum_{p \in \Omega}
\left(u_p^{T}w\right)^2
=
w^{T}\Sigma_{\Omega}w,
\qquad
\Sigma_{\Omega}
=
\frac{1}{|\Omega|-1}
\sum_{p \in \Omega}
u_pu_p^{T}.
\label{eq:p3ca_local_contrast}
\end{equation}
Thus, maximizing local contrast corresponds to finding dominant directions of the region-conditioned covariance matrix. For an RGB representation, we seek three orthonormal directions that jointly maximize the projected variance:
\begin{equation}
W_{\Omega}
=
\arg\max_{W^{T}W=I_3}
\operatorname{tr}(W^{T}\Sigma_{\Omega}W)
=
[w_1,w_2,w_3]
\in \mathbb{R}^{C \times 3}.
\label{eq:p3ca_basis}
\end{equation}
The solution is given by the top three eigenvectors of $\Sigma_{\Omega}$, showing that the local-contrast objective leads to PCA fitted within the prompted region. We refer to this procedure as position-prompted PCA. The locally estimated basis is then applied to every location in the full tensor,
\begin{equation}
z_p = W_{\Omega}^{T}u_p \in \mathbb{R}^{3},
\qquad p \in \mathcal{G},
\label{eq:p3ca_projection}
\end{equation}
producing $Z_{\Omega} \in \mathbb{R}^{H \times W \times 3}$, which is normalized only for RGB display. Global PCA is recovered when $\Omega=\mathcal{G}$; otherwise, P\textsuperscript{3}CA visualizes where the directions that maximize local contrast in the prompt are expressed across the image.

For quantitative comparison, we define regional contrast gain using the frozen numerical projection coordinates before RGB display normalization. For an evaluation region $R$ and method $m \in \{\mathcal{G},\Omega\}$, let $F_m$ denote the fitted normalization and projection pipeline. We compute
\begin{equation}
V_m(R)=\operatorname{tr}\!\left(\operatorname{Cov}(F_m(X_R))\right),
\qquad
CG_{\Omega}(R)=\frac{V_{\Omega}(R)}{V_{\mathcal{G}}(R)} .
\label{eq:p3ca_contrast_gain}
\end{equation}
Values above one indicate that P\textsuperscript{3}CA preserves more three-dimensional feature variation in $R$ than the global projection.

\subsection{Interactive workflow and implementation}

\begin{figure}[!t]
\centering
\includegraphics[width=0.8\textwidth]{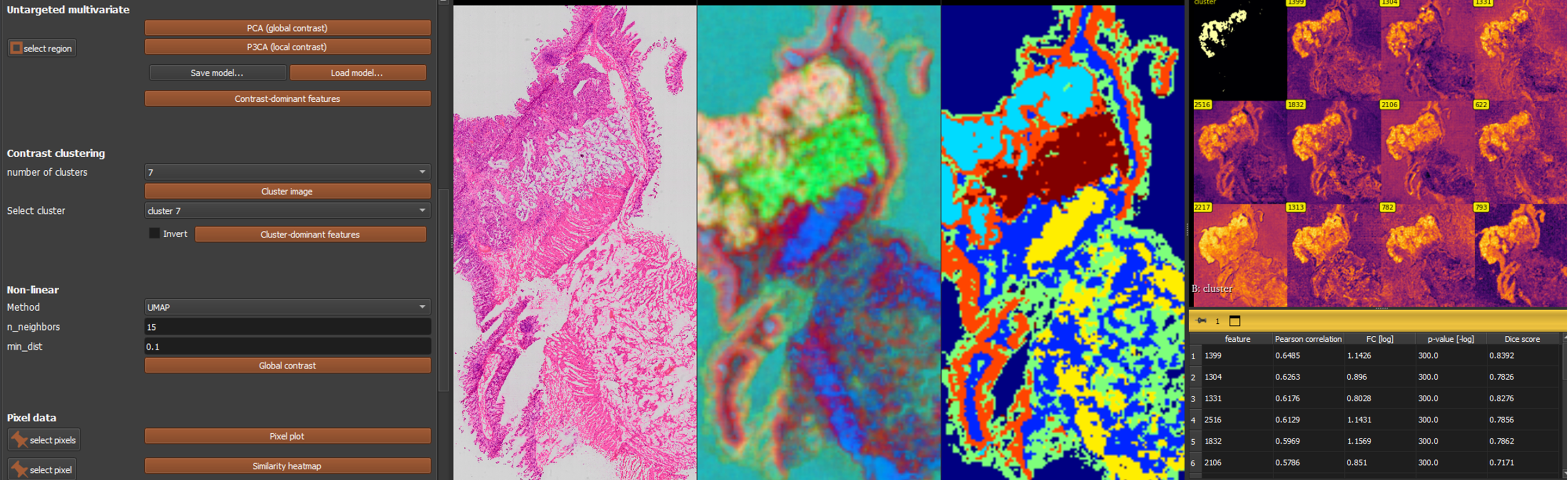}
\caption{EmbedVision interface for spatial representation analysis, showing prompt-based projections, clustering, overlays, and dominant-feature summaries side by side.}
\label{fig:embedvision}
\end{figure}

We implemented P\textsuperscript{3}CA in EmbedVision\footnote{\url{https://slicermassvision.readthedocs.io/en/latest/EmbedVision.html}} (Fig.~\ref{fig:embedvision}), an interactive Python/Qt workflow built on the open-source 3D Slicer~\cite{slicer} and MassVision~\cite{massvision} platforms. EmbedVision loads a spatial tensor and optional reference image, supports rectangular or irregular prompts, computes global PCA or P\textsuperscript{3}CA, and displays the resulting representation map with image overlays. Because P\textsuperscript{3}CA is prompt-driven, the interface is designed for iterative inspection: users can move or redraw prompts, compare the resulting local projections with the global projection, and inspect whether prompt-relevant structure appears elsewhere in the image.

To support reproducible comparison, EmbedVision stores the fitted preprocessing and projection transforms together with the prompt mask and display parameters. These transforms can be reapplied to tensors from the same feature space, enabling cross-image or cross-sample comparison without refitting the projection. The same workflow supports learned embedding tensors and measured spatial tensors, allowing foundation-model features, molecular measurements, and gene-expression data to be explored within a common spatial interface. The interface also includes UMAP, t-SNE, similarity heatmaps, clustering maps, channel-contribution summaries, and overlay tools for comparing prompts, encoders, and modalities. 

\subsection{Experiments and Case Studies}

We evaluated P\textsuperscript{3}CA in three complementary settings designed to test different forms of region-aware representation interpretation. First, a natural-image example encoded with DINOv3~\cite{simeoni2025dinov3} was used as a baseline to assess whether spatial prompts expose local anatomical structure that is not emphasized by global PCA. Second, a colorectal whole-slide image~\cite{colon23} was processed with two pathology foundation-model encoders, GigaPath~\cite{xu2024gigapath} and H0-mini~\cite{filiot2025h0mini}, to evaluate prompt-specific tissue contrast and downstream classification from the resulting projections. Finally, a glioblastoma sample~\cite{ravi2022data} with co-registered H\&E and 10x Visium spatial transcriptomics was used to compare morphology-derived CONCH embeddings~\cite{lu2024conch} with native gene-expression tensors.

Prompts were selected from spatially meaningful regions in each study. Readouts were matched to the purpose of each case: regional contrast gain and annotated patch scatter plots for natural-image and pathology examples, LDA balanced accuracy for pathology, and agreement with Greenwald module assignments~\cite{greenwald2024integrative} as a post hoc biological reference in spatial transcriptomics. Prompts and biological labels were not used to train or modify the encoders.

\section{Results and Discussion}

\begin{figure}[!t]
\centering
\includegraphics[width=0.9\textwidth]{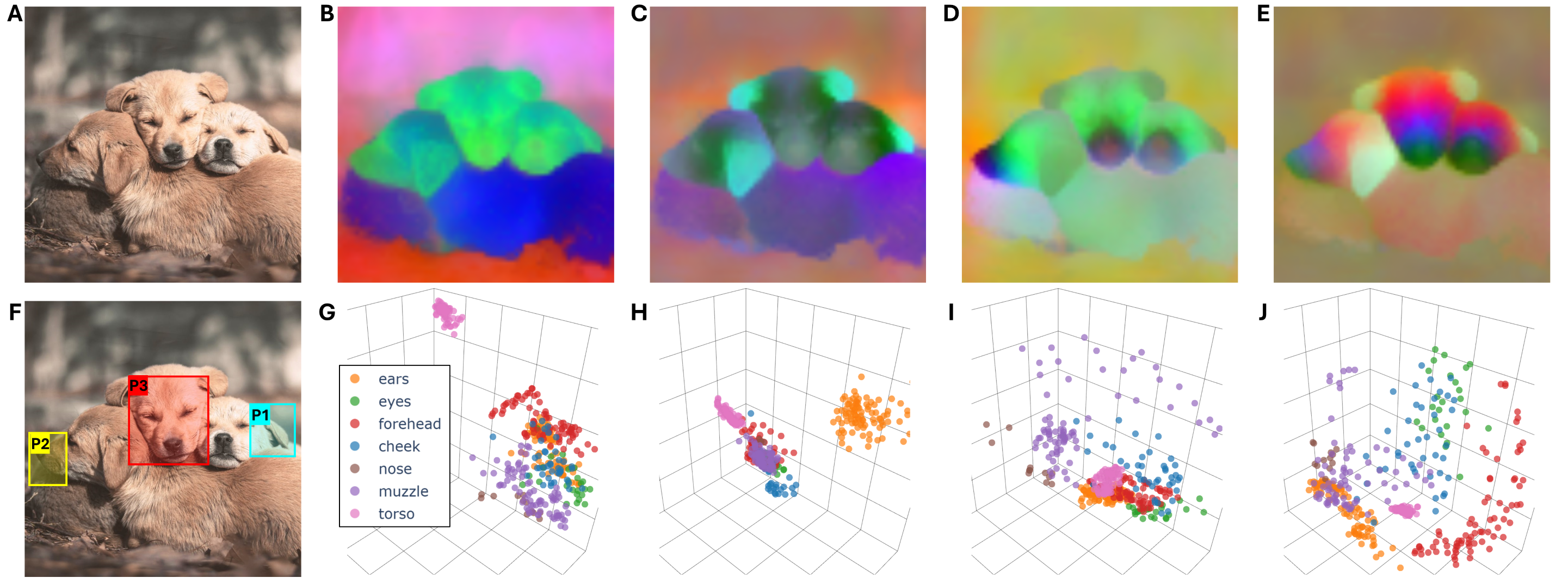}
\caption{Natural-image evaluation. (A) Input image; (B) global PCA; (C--E) P\textsuperscript{3}CA maps from prompts in (F). (G--J) 3D scatter plots of a manually annotated subset of patches, obtained by applying the corresponding projections from (B--E) to the labeled data.}
\label{fig:natural}
\end{figure}

\noindent \textbf{Local visual probing in natural images:} 
Figure~\ref{fig:natural} evaluates P\textsuperscript{3}CA on a natural image with a manually annotated subset of patches, colored by anatomical labels in the 3D projection spaces (Fig.~\ref{fig:natural}G--J). Global PCA captures coarse separation between the puppies and background, but facial subregions remain mixed in both the RGB map and projection space (Fig.~\ref{fig:natural}B,G). In contrast, prompted projections selectively increase variation associated with the selected local structure: the ear, muzzle/nose, and broader face prompts achieve contrast gains of 8.39, 11.26, and 13.56 over global PCA, respectively. All three gains exceed one, indicating higher local contrast than the global projection.

Applying the same projections to the labeled patches shows that ear and muzzle/nose prompts better separate their corresponding facial structures, while the broader face prompt gives the clearest differentiation among facial regions (Fig.~\ref{fig:natural}H--J). This controlled example establishes the basic behavior of P\textsuperscript{3}CA: DINOv3 embeddings contain fine-grained local structure that is suppressed by a global projection but can be exposed by changing only the spatial region used to fit the projection. It also illustrates an expected trade-off: optimizing contrast for a local prompt can compress unrelated regions, because P\textsuperscript{3}CA is a region-conditioned representation lens rather than a globally balanced visualization.\\

\begin{figure}[!t]
\centering
\includegraphics[width=0.9\textwidth]{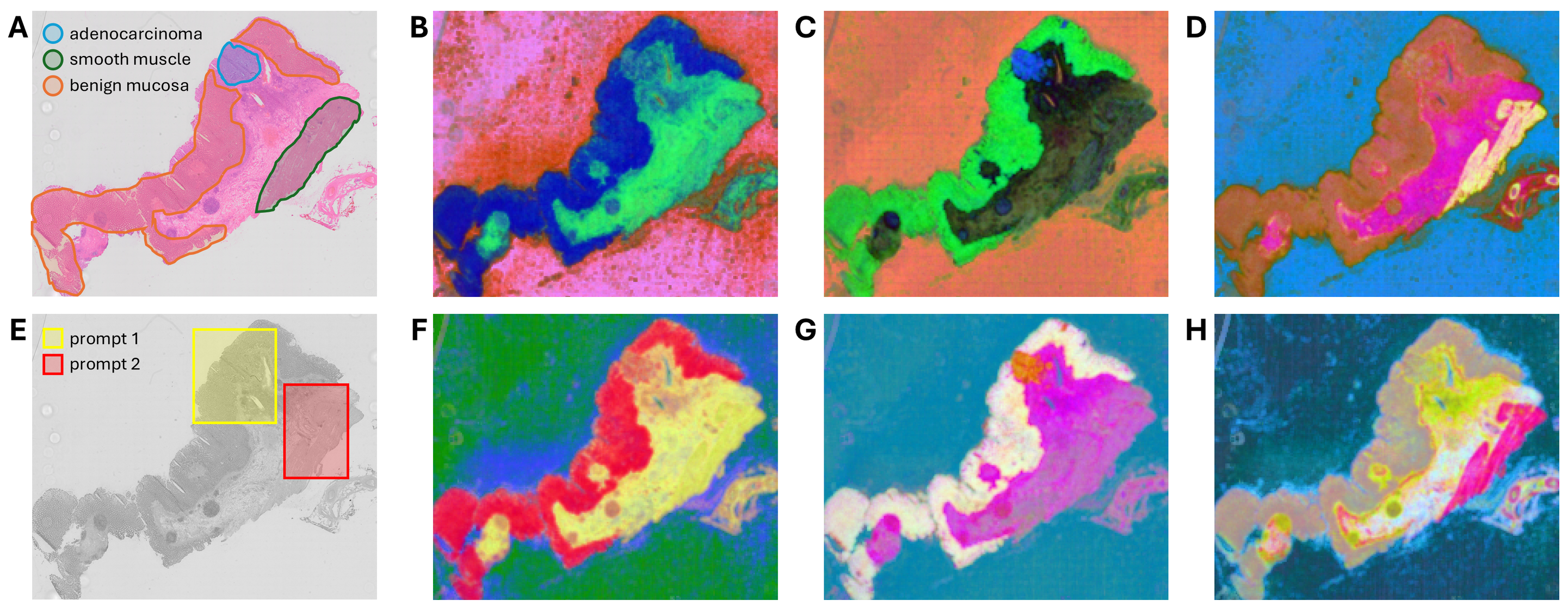}
\caption{Pathology foundation-model evaluation. (A) Colorectal whole-slide image with expert annotations; (E) spatial prompts. (B,F) Global PCA projections for GigaPath and H0-mini. P\textsuperscript{3}CA projections associated with prompts 1 and 2 for GigaPath (C,D) and H0-mini (G,H).}
\label{fig:pathology}
\end{figure}

\noindent\textbf{Prompt-specific pathology contrast and downstream classification:}
Figure~\ref{fig:pathology} evaluates P\textsuperscript{3}CA on colorectal pathology embeddings from GigaPath and H0-mini. The whole-slide image includes expert tissue annotations, while two spatial prompts were selected for local projection analysis (Fig.~\ref{fig:pathology}A,E). Global PCA captures broad tissue-level variation but merges several diagnostically relevant regions. 
In contrast, P\textsuperscript{3}CA projections associated with prompts 1 and 2 produce stronger local contrast in their selected regions for both encoders, with contrast gains of 2.48 and 2.24 for GigaPath, and 2.31 and 2.01 for H0-mini, respectively. Qualitatively, the prompted projections also make the corresponding annotated tissue structures more visually distinguishable than the global projections.

We next tested whether the same frozen three-dimensional projections used for visualization also preserve information useful for downstream discrimination. Patch-level annotations from nine patients were projected using the PCA/P\textsuperscript{3}CA transforms fitted as in Fig.~\ref{fig:pathology}; these transforms were then kept fixed throughout leave-one-patient-out evaluation. For each fold, only an LDA classifier was trained, using the three projected coordinates as input features, and balanced accuracy was evaluated on the held-out patient.

As presented in Table~\ref{tab:pathology_classification}, prompt 1 significantly improved adenocarcinoma classification to 85.71\% for GigaPath and 85.41\% for H0-mini, compared with 69.07\% and 59.01\% for global PCA (Wilcoxon signed-rank test, $p<0.01$). Prompt 2 significantly improved smooth-muscle classification to 84.04\% and 79.50\%, compared with 49.19\% and 53.63\% for global PCA ($p<0.01$). Non-matching prompts did not provide comparable gains. Thus, the same prompt-conditioned projections that improve visual contrast also define fixed 3D representations that remain discriminative for tissue classification.
\\

\begin{table}[t]
\centering
\caption{Balanced accuracy (\%) for LDA classification using frozen three-dimensional pathology projections and leave-one-patient-out cross-validation. PCA/P\textsuperscript{3}CA transforms were fitted as in Fig.~\ref{fig:pathology}.}
\label{tab:pathology_classification}
\small
\fontsize{8pt}{10pt}\selectfont
\setlength{\tabcolsep}{4pt}
\begin{tabular}{lcccc}
\toprule
 & \multicolumn{2}{c}{\textbf{Adenocarcinoma}} 
 & \multicolumn{2}{c}{\textbf{Smooth muscle}} \\
\cmidrule(lr){2-3} \cmidrule(lr){4-5}
\textbf{Projection} 
& \textbf{GigaPath} & \textbf{H0-mini} 
& \textbf{GigaPath} & \textbf{H0-mini} \\
\midrule
PCA 
& $69.07 \pm 16.23$ 
& $59.01 \pm 12.66$ 
& $49.19 \pm 0.90$ 
& $53.63 \pm 7.15$ \\
P\textsuperscript{3}CA prompt 1 
& $\mathbf{85.71 \pm 11.97}$ 
& $\mathbf{85.41 \pm 9.22}$ 
& $48.35 \pm 2.33$ 
& $49.32 \pm 3.12$ \\
P\textsuperscript{3}CA prompt 2 
& $63.73 \pm 11.42$ 
& $63.63 \pm 15.11$ 
& $\mathbf{84.04 \pm 14.92}$ 
& $\mathbf{79.50 \pm 12.43}$ \\
\bottomrule
\end{tabular}
\end{table}

\begin{figure}[!t]
\centering
\includegraphics[width=\textwidth]{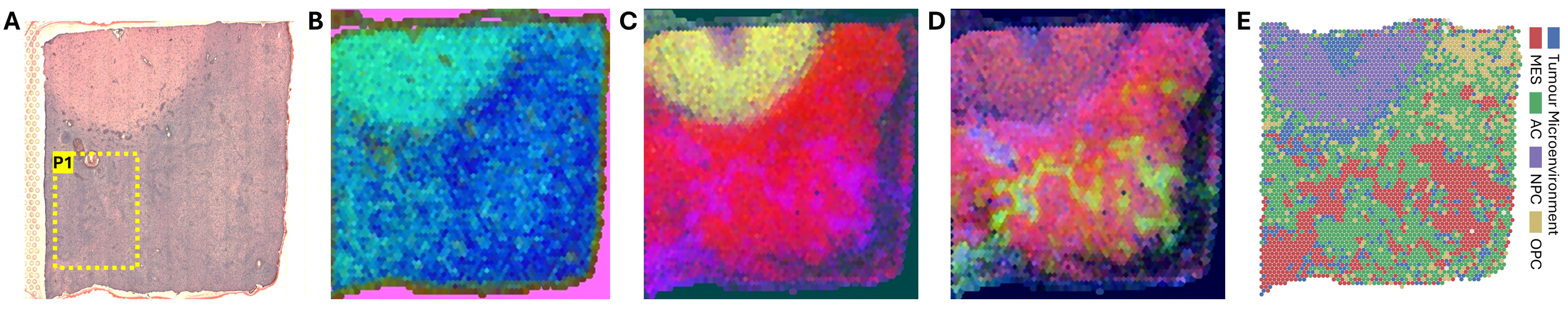}
\caption{Multimodal glioblastoma analysis. (A) H\&E section with spatial prompt; (B) CONCH pathology PCA; (C) transcriptomic PCA; (D) transcriptomic P\textsuperscript{3}CA for prompt in (A); (E) Greenwald module reference.}
\label{fig:multimodal}
\end{figure}

\noindent\textbf{Multimodal discovery in spatial transcriptomics:}
Figure~\ref{fig:multimodal} applies the same spatial probing workflow to a glioblastoma sample with co-registered H\&E morphology and 10x Visium transcriptomics. We compared a morphology-derived CONCH embedding with the native gene-expression tensor, using Greenwald module assignments only as a post hoc biological reference and not during PCA or P\textsuperscript{3}CA fitting. Global PCA of the CONCH embedding captures broad morphology-associated variation, but does not clearly recover the microenvironment organization shown by the module reference map. In contrast, global PCA of the transcriptomic tensor better reflects the major molecular compartments.

P\textsuperscript{3}CA further refines this view by emphasizing variation within the prompted region. The transcriptomic P\textsuperscript{3}CA map reveals local heterogeneity around the malignant region, including a boundary between malignant transcriptional programs and the surrounding non-malignant tumour microenvironment that is less apparent in the global projection. This example extends P\textsuperscript{3}CA beyond learned image embeddings: the same local-contrast framework can probe measured molecular tensors and compare them with morphology-derived foundation-model features in a shared spatial coordinate system.\\

\noindent\textbf{Limitations:}
P\textsuperscript{3}CA is intended as an exploratory representation-probing tool rather than a standalone validation metric. Its output depends on the selected prompt, and optimizing contrast for one region can compress unrelated structures elsewhere in the image. The projection is linear, so nonlinear relationships in the embedding space may not be fully captured by three components. In addition, RGB colors should be interpreted as coordinates in a fitted projection space rather than fixed semantic labels. Larger studies are needed to evaluate prompt stability, inter-user variability, and the relationship between local projection structure and downstream clinical performance.

\section{Conclusions}

We introduced P\textsuperscript{3}CA, an encoder-agnostic method for region-aware interpretation of high-dimensional spatial representations. By fitting a local-contrast projection within a spatial prompt and applying the resulting basis to the full tensor, P\textsuperscript{3}CA provides a representation lens that complements global visualization, downstream probes, and attribution methods. Across natural images, pathology foundation-model embeddings, and spatial transcriptomics, the method exposed locally meaningful structure suppressed by global PCA and supported qualitative inspection, quantitative contrast analysis, and prompt-matched classification. Future work will evaluate larger cohorts, additional foundation models, and systematic prompt-selection strategies.

 %% removed for anonymized MICCAI submission.
    
    % The following acknowledgement and disclaimer sections can be removed for the double-blind review process.  If and when your paper is accepted, reinsert the acknowledgement and the disclaimer clause in your final camera-ready version.
    % IF you opted to include the acknowledgement and disclaimer sections, they will count towards the 8-page limit.

\begin{credits}
\subsubsection{\ackname} 
The project is supported in part by NSERC and CIHR. Parvin Mousavi is supported by a Canada Research Chair in Medical Informatics, Canada CIFAR AI Chair, and the Vector Institute.

\subsubsection{\discintname}
The authors have no competing interests to declare.
\end{credits}

%
% ---- Bibliography ----
%
% BibTeX users should specify bibliography style 'splncs04'.
% References will then be sorted and formatted in the correct style.

\bibliographystyle{splncs04}
\bibliography{main-biblio}

\end{document}